\documentclass[letterpaper]{article} 
\usepackage{aaai24}  
\usepackage{times}  
\usepackage{helvet}  
\usepackage{courier}  
\usepackage[hyphens]{url}  
\usepackage{graphicx} 
\usepackage{natbib}  
\usepackage{caption} 
\usepackage{algorithm}

\usepackage{newfloat}
\usepackage{listings}
\DeclareCaptionStyle{ruled}{labelfont=normalfont,labelsep=colon,strut=off} 
\floatstyle{ruled}
\newfloat{listing}{tb}{lst}{}
\floatname{listing}{Listing}
\usepackage[T1]{fontenc}
\usepackage[utf8]{inputenc}
\usepackage{mathtools}

\usepackage{amssymb,mathrsfs}
\usepackage{amsthm}
\usepackage{bm}
\usepackage{scalerel}
\usepackage{nicefrac}
\usepackage{microtype} 
\usepackage[shortlabels]{enumitem}
\usepackage{graphicx}
\usepackage{epstopdf}
\DeclareGraphicsExtensions{.eps,.png,.jpg,.pdf}

\usepackage{url}
\usepackage{colortbl}
\usepackage{booktabs}
\usepackage{multirow}
\usepackage{colortbl,xcolor}
\usepackage{xparse,xstring}
\usepackage{calc}
\usepackage{etoolbox}

\makeatletter
\@ifpackageloaded{natbib}{
	\relax
}{
	\usepackage{cite}
}
\makeatother

\usepackage{array}
\newcolumntype{L}[1]{>{\raggedright\let\newline\\\arraybackslash\hspace{0pt}}m{#1}}
\newcolumntype{C}[1]{>{\centering\let\newline\\\arraybackslash\hspace{0pt}}m{#1}}
\newcolumntype{R}[1]{>{\raggedleft\let\newline\\\arraybackslash\hspace{0pt}}m{#1}}

\makeatletter
\let\MYcaption\@makecaption
\makeatother
\usepackage[font=footnotesize]{subcaption}
\makeatletter
\let\@makecaption\MYcaption
\makeatother

\usepackage{glossaries}
\makeatletter
\sfcode`\.1006

\let\oldgls\gls
\let\oldglspl\glspl

\newcommand\fussy@ifnextchar[3]{%
	\let\reserved@d=#1%
	\def\reserved@a{#2}%
	\def\reserved@b{#3}%
	\futurelet\@let@token\fussy@ifnch}
\def\fussy@ifnch{%
	\ifx\@let@token\reserved@d
		\let\reserved@c\reserved@a
	\else
		\let\reserved@c\reserved@b
	\fi
	\reserved@c}

\renewcommand{\gls}[1]{%
\oldgls{#1}\fussy@ifnextchar.{\@checkperiod}{\@}}
\renewcommand{\glspl}[1]{%
\oldglspl{#1}\fussy@ifnextchar.{\@checkperiod}{\@}}

\newcommand{\@checkperiod}[1]{%
	\ifnum\sfcode`\.=\spacefactor\else#1\fi
}

\robustify{\gls}
\robustify{\glspl}
\makeatother

\newacronym{wrt}{w.r.t.}{with respect to}
\newacronym{RHS}{R.H.S.}{right-hand side}
\newacronym{LHS}{L.H.S.}{left-hand side}
\newacronym{iid}{i.i.d.}{independent and identically distributed}
\newacronym{SOTA}{SOTA}{state-of-the-art}

\usepackage{float}

\usepackage[capitalize]{cleveref}
\crefname{equation}{}{}
\Crefname{equation}{}{}
\crefname{claim}{claim}{claims}
\crefname{step}{step}{steps}
\crefname{line}{line}{lines}
\crefname{condition}{condition}{conditions}
\crefname{dmath}{}{}
\crefname{dseries}{}{}
\crefname{dgroup}{}{}

\crefname{Problem}{Problem}{Problems}
\crefformat{Problem}{Problem~#2#1#3}
\crefrangeformat{Problem}{Problems~#3#1#4 to~#5#2#6}

\crefname{Theorem}{Theorem}{Theorems}
\crefname{Corollary}{Corollary}{Corollaries}
\crefname{Proposition}{Proposition}{Propositions}
\crefname{Lemma}{Lemma}{Lemmas}
\crefname{Definition}{Definition}{Definitions}
\crefname{Example}{Example}{Examples}
\crefname{Assumption}{Assumption}{Assumptions}
\crefname{Remark}{Remark}{Remarks}
\crefname{Rem}{Remark}{Remarks}
\crefname{remarks}{Remarks}{Remarks}
\crefname{Appendix}{Appendix}{Appendices}
\crefname{Supplement}{Supplement}{Supplements}
\crefname{Exercise}{Exercise}{Exercises}
\crefname{Theorem_A}{Theorem}{Theorems}
\crefname{Corollary_A}{Corollary}{Corollaries}
\crefname{Proposition_A}{Proposition}{Propositions}
\crefname{Lemma_A}{Lemma}{Lemmas}
\crefname{Definition_A}{Definition}{Definitions}

\usepackage{crossreftools}

\usepackage{algorithm}
\usepackage{algpseudocode}

\makeatletter
\def\cleartheorem#1{%
    \expandafter\let\csname#1\endcsname\relax
    \expandafter\let\csname c@#1\endcsname\relax
}
\def\clearthms#1{ \@for\tname:=#1\do{\cleartheorem\tname} }
\makeatother

\ifx\renewtheorem\undefined
	\ifx\useTheoremCounter\undefined
		\newtheorem{Theorem}{Theorem}
		\newtheorem{Corollary}{Corollary}
		\newtheorem{Proposition}{Proposition}
		
	\else

	\fi

\fi

\theoremstyle{remark}

\theoremstyle{plain}

\newcommand{\qednew}{\nobreak \ifvmode \relax \else
		\ifdim\lastskip<1.5em \hskip-\lastskip
			\hskip1.5em plus0em minus0.5em \fi \nobreak
		\vrule height0.75em width0.5em depth0.25em\fi}



\NewDocumentCommand{\movedownsub}{e{^_}}{%
	\IfNoValueTF{#1}{%
		\IfNoValueF{#2}{^{}}
	}{%
		^{#1}
	}%
	\IfNoValueF{#2}{_{#2}}
}

\let\latexchi\chi
\RenewDocumentCommand{\chi}{}{\latexchi\movedownsub}

\newcommand{\calL}{\mathcal{L}}
\newcommand{\calM}{\mathcal{M}}

\newcommand{\bg}{\mathbf{g}}

\newcommand{\bI}{\mathbf{I}}

\newcommand{\bp}{\mathbf{p}}

\newcommand{\bq}{\mathbf{q}}

\newcommand{\bs}{\mathbf{s}}
\newcommand{\bS}{\mathbf{S}}
\newcommand{\bt}{\mathbf{t}}
\newcommand{\bT}{\mathbf{T}}

\newcommand{\bv}{\mathbf{v}}

\newcommand{\bx}{\mathbf{x}}

\newcommand{\by}{\mathbf{y}}

\newcommand{\bz}{\mathbf{z}}

\newcommand{\bbD}{\mathbb{D}}

\newcommand{\bbR}{\mathbb{R}}

\DeclareSymbolFont{bsfletters}{OT1}{cmss}{bx}{n}
\DeclareSymbolFont{ssfletters}{OT1}{cmss}{m}{n}
\DeclareMathSymbol{\bsfGamma}{0}{bsfletters}{'000}
\DeclareMathSymbol{\ssfGamma}{0}{ssfletters}{'000}
\DeclareMathSymbol{\bsfDelta}{0}{bsfletters}{'001}
\DeclareMathSymbol{\ssfDelta}{0}{ssfletters}{'001}
\DeclareMathSymbol{\bsfTheta}{0}{bsfletters}{'002}
\DeclareMathSymbol{\ssfTheta}{0}{ssfletters}{'002}
\DeclareMathSymbol{\bsfLambda}{0}{bsfletters}{'003}
\DeclareMathSymbol{\ssfLambda}{0}{ssfletters}{'003}
\DeclareMathSymbol{\bsfXi}{0}{bsfletters}{'004}
\DeclareMathSymbol{\ssfXi}{0}{ssfletters}{'004}
\DeclareMathSymbol{\bsfPi}{0}{bsfletters}{'005}
\DeclareMathSymbol{\ssfPi}{0}{ssfletters}{'005}
\DeclareMathSymbol{\bsfSigma}{0}{bsfletters}{'006}
\DeclareMathSymbol{\ssfSigma}{0}{ssfletters}{'006}
\DeclareMathSymbol{\bsfUpsilon}{0}{bsfletters}{'007}
\DeclareMathSymbol{\ssfUpsilon}{0}{ssfletters}{'007}
\DeclareMathSymbol{\bsfPhi}{0}{bsfletters}{'010}
\DeclareMathSymbol{\ssfPhi}{0}{ssfletters}{'010}
\DeclareMathSymbol{\bsfPsi}{0}{bsfletters}{'011}
\DeclareMathSymbol{\ssfPsi}{0}{ssfletters}{'011}
\DeclareMathSymbol{\bsfOmega}{0}{bsfletters}{'012}
\DeclareMathSymbol{\ssfOmega}{0}{ssfletters}{'012}

\makeatletter
\newcommand*\rel@kern[1]{\kern#1\dimexpr\macc@kerna}
\newcommand*\widebar[1]{%
  \begingroup
  \def\mathaccent##1##2{%
    \rel@kern{0.8}%
    \overline{\rel@kern{-0.8}\macc@nucleus\rel@kern{0.2}}%
    \rel@kern{-0.2}%
  }%
  \macc@depth\@ne
  \let\math@bgroup\@empty \let\math@egroup\macc@set@skewchar
  \mathsurround\z@ \frozen@everymath{\mathgroup\macc@group\relax}%
  \macc@set@skewchar\relax
  \let\mathaccentV\macc@nested@a
  \macc@nested@a\relax111{#1}%
  \endgroup
}
\makeatother

\DeclareMathOperator{\var}{var}

\DeclareMathOperator{\cov}{cov}

\newcommand{\ifbcdot}[1]{\ifblank{#1}{\cdot}{#1}}

\DeclarePairedDelimiterX\abs[1]{\lvert}{\rvert}{\ifbcdot{#1}}
\DeclarePairedDelimiterX\parens[1]{(}{)}{\ifbcdot{#1}}
\DeclarePairedDelimiterX\brk[1]{[}{]}{\ifbcdot{#1}}
\DeclarePairedDelimiterX\braces[1]{\{}{\}}{\ifbcdot{#1}}
\DeclarePairedDelimiterX\angles[1]{\langle}{\rangle}{\ifblank{#1}{\cdot,\cdot}{#1}}
\DeclarePairedDelimiterX\ip[2]{\langle}{\rangle}{\ifbcdot{#1},\ifbcdot{#2}}
\DeclarePairedDelimiterX\norm[1]{\lVert}{\rVert}{\ifbcdot{#1}}
\DeclarePairedDelimiterX\ceil[1]{\lceil}{\rceil}{\ifbcdot{#1}}
\DeclarePairedDelimiterX\floor[1]{\lfloor}{\rfloor}{\ifbcdot{#1}}

\DeclareFontFamily{U}{matha}{\hyphenchar\font45}
\DeclareFontShape{U}{matha}{m}{n}{
      <5> <6> <7> <8> <9> <10> gen * matha
      <10.95> matha10 <12> <14.4> <17.28> <20.74> <24.88> matha12
      }{}
\DeclareSymbolFont{matha}{U}{matha}{m}{n}
\DeclareFontSubstitution{U}{matha}{m}{n}

\DeclareFontFamily{U}{mathx}{\hyphenchar\font45}
\DeclareFontShape{U}{mathx}{m}{n}{
      <5> <6> <7> <8> <9> <10>
      <10.95> <12> <14.4> <17.28> <20.74> <24.88>
      mathx10
      }{}
\DeclareSymbolFont{mathx}{U}{mathx}{m}{n}
\DeclareFontSubstitution{U}{mathx}{m}{n}

\DeclareMathDelimiter{\vvvert}{0}{matha}{"7E}{mathx}{"17}
\DeclarePairedDelimiterX\vertiii[1]{\vvvert}{\vvvert}{\ifbcdot{#1}}

\DeclarePairedDelimiterXPP\trace[1]{\operatorname{Tr}}{(}{)}{}{\ifbcdot{#1}} 
\DeclarePairedDelimiterXPP\col[1]{\operatorname{col}}{\{}{\}}{}{\ifbcdot{#1}} 
\DeclarePairedDelimiterXPP\row[1]{\operatorname{row}}{\{}{\}}{}{\ifbcdot{#1}} 
\DeclarePairedDelimiterXPP\erf[1]{\operatorname{erf}}{(}{)}{}{\ifbcdot{#1}}
\DeclarePairedDelimiterXPP\erfc[1]{\operatorname{erfc}}{(}{)}{}{\ifbcdot{#1}}
\DeclarePairedDelimiterXPP\KLD[2]{D}{(}{)}{}{\ifbcdot{#1}\, \delimsize\|\, \ifbcdot{#2}} 
\DeclarePairedDelimiterXPP\op[2]{\operatorname{#1}}{(}{)}{}{#2} 

\DeclarePairedDelimiterXPP\indicate[1]{{\bf 1}}{\{}{\}}{}{\ifbcdot{#1}}

\NewDocumentCommand\ofrac{s m}{%
	\IfBooleanTF#1%
	{\dfrac{1}{#2}}%
	{\frac{1}{#2}}%
}
\NewDocumentCommand\ddfrac{s m m}{%
	\IfBooleanTF#1%
	{\dfrac{\mathrm{d} {#2}}{\mathrm{d} {#3}}}%
	{\frac{\mathrm{d} {#2}}{\mathrm{d} {#3}}}%
}
\NewDocumentCommand\ppfrac{s m m}{%
	\IfBooleanTF#1%
	{\dfrac{\partial {#2}}{\partial {#3}}}%
	{\frac{\partial {#2}}{\partial {#3}}}%
}

\providecommand\given{}

\DeclarePairedDelimiterX\Set[2]\{\}{%
\renewcommand\given{\SetSymbol[\delimsize]{#1}}
#2
}
\DeclarePairedDelimiterX\Setc[1]\{\}{%
\renewcommand\given{\SetSymbol{:}}
#1
}

\NewDocumentCommand\set{s o m}{%
	\IfBooleanTF#1%
	{\IfValueTF{#2}{\Set*{#2}{#3}}{\Setc*{#3}}}%
	{\IfValueTF{#2}{\Set{#2}{#3}}{\Setc{#3}}}%
}

\NewDocumentCommand{\evalat}{ s O{\big} m e{_^} }{%
\IfBooleanTF{#1}%
{\left. #3 \right|}{#3#2|}%
\IfValueT{#4}{_{#4}}%
\IfValueT{#5}{^{#5}}%
}

\providecommand\given{}
\DeclarePairedDelimiterXPP\cprob[1]{}(){}{
\renewcommand\given{\nonscript\,\delimsize\vert\allowbreak\nonscript\,\mathopen{}}%
#1%
}
\DeclarePairedDelimiterXPP\cexp[1]{}[]{}{
\renewcommand\given{\nonscript\,\delimsize\vert\allowbreak\nonscript\,\mathopen{}}%
#1%
}

\DeclareDocumentCommand \P { s e{_^} d() g } {%
	\mathbb{P}%
	\IfBooleanTF{#1}%
		{
			\IfValueT{#2}{_{#2}}%
			\IfValueT{#3}{^{#3}}%
			\IfValueTF{#5}{\cprob{#4 \given #5}}{\IfValueT{#4}{\cprob{#4}}}%
		}%
		{
			\IfValueT{#2}{_{#2}}%
			\IfValueT{#3}{^{#3}}%
			\IfValueTF{#5}{\cprob*{#4 \given #5}}{\IfValueT{#4}{\cprob*{#4}}}%
		}%
}

\DeclareDocumentCommand \E { s e{_^} o g } {%
	\mathbb{E}%
	\IfBooleanTF{#1}%
		{
			\IfValueT{#2}{_{#2}}%
			\IfValueT{#3}{^{#3}}%
			\IfValueTF{#5}{\cexp{#4 \given #5}}{\IfValueT{#4}{\cexp{#4}}}%
		}%
		{
			\IfValueT{#2}{_{#2}}%
			\IfValueT{#3}{^{#3}}%
			\IfValueTF{#5}{\cexp*{#4 \given #5}}{\IfValueT{#4}{\cexp*{#4}}}%
		}%
}

\DeclareDocumentCommand \Var { s e{_^} d() g } {%
	\var%
	\IfBooleanTF{#1}%
		{
			\IfValueT{#2}{_{#2}}%
			\IfValueT{#3}{^{#3}}%
			\IfValueTF{#5}{\cprob{#4 \given #5}}{\IfValueT{#4}{\cprob{#4}}}%
		}%
		{
			\IfValueT{#2}{_{#2}}%
			\IfValueT{#3}{^{#3}}%
			\IfValueTF{#5}{\cprob*{#4 \given #5}}{\IfValueT{#4}{\cprob*{#4}}}%
		}%
}

\DeclareDocumentCommand \Cov { s e{_^} d() g } {%
	\cov%
	\IfBooleanTF{#1}%
		{
			\IfValueT{#2}{_{#2}}%
			\IfValueT{#3}{^{#3}}%
			\IfValueTF{#5}{\cprob{#4 \given #5}}{\IfValueT{#4}{\cprob{#4}}}%
		}%
		{
			\IfValueT{#2}{_{#2}}%
			\IfValueT{#3}{^{#3}}%
			\IfValueTF{#5}{\cprob*{#4 \given #5}}{\IfValueT{#4}{\cprob*{#4}}}%
		}%
}

\ExplSyntaxOn
\NewDocumentCommand \dist {m o o} {%
\mathrm{#1}\left(%
	\IfValueT{#3}{%
		\tl_if_blank:nTF{ #3 }{\cdot\, \middle|\, }{#3\, \middle|\, }%
	}
	\IfValueT{#2}{#2}%
\right)%
}
\ExplSyntaxOff

\NewDocumentCommand {\cbrace} {t+ D[]{black} D(){\widthof{#5}} m m } {%
	\begingroup%
		\color{#2}
		\IfBooleanTF{#1}{%
			\overbrace{#4}^%
		}{
			\underbrace{#4}_%
		}%
		{\parbox[c]{#3}{\centering\footnotesize{#5}}}%
	\endgroup%
}

\let\oldforall\forall
\renewcommand{\forall}{\oldforall \, }

\let\oldexist\exists
\renewcommand{\exists}{\oldexist \, }

\makeatletter

\newcommand{\rankcolor}[2]{%
	\expandafter\renewcommand\csname #1\endcsname[1]{%
		\ifblank{##1}{%
			{\color{#2} \textbf{#2}}%
		}{%
			\ifmmode
				\textcolor{#2}{\bm{##1}}%
			\else%
				{\color{#2} \textbf{##1}}%
			\fi	
		}%
	}
}

\rankcolor{first}{red}
\rankcolor{second}{blue}
\rankcolor{third}{cyan}
\makeatother

\graphicspath{{./Figures/}{./figures/}}
\DeclareDocumentCommand{\includeCroppedPdf}{ o O{./Figures/} m }{
	\IfFileExists{#2#3-crop.pdf}{}{%
		\immediate\write18{pdfcrop #2#3.pdf #2#3-crop.pdf}}%
	\includegraphics[#1]{#2#3-crop.pdf}
}

\makeatletter
\newcommand*{\addFileDependency}[1]{
  \typeout{(#1)}
  \@addtofilelist{#1}
  \IfFileExists{#1}{}{\typeout{No file #1.}}
}
\makeatother

\definecolor{gray90}{gray}{0.9}
\def\colorlist{red,blue,brown,cyan,darkgray,gray,lightgray,green,lime,magenta,olive,orange,pink,purple,teal,violet,white,yellow}

\makeatletter
\def\startcomment{[}
\ifx\nohighlights\undefined
	\newcommand{\createcolor}[1]{%
			\expandafter\newcommand\csname #1\endcsname[1]{{\color{#1} ##1}}%
	}
	\newcommand{\msout}[1]{\text{\color{green} \sout{\ensuremath{#1}}}}
	\newcommand{\del}[1]{{\color{green}\ifmmode \msout{#1}\else\sout{#1}\fi}}
\else
	\newcommand{\createcolor}[1]{%
			\expandafter\newcommand\csname #1\endcsname[1]{%
				\noexpandarg%
				\StrChar{##1}{1}[\firstletter]%
				\if\firstletter\startcomment%
					\relax
				\else%
					##1
				\fi
			}%
	}
	\newcommand{\msout}[1]{}
	\newcommand{\del}[1]{}
\fi

\def\@tempa#1,{%
    \ifx\relax#1\relax\else
        \createcolor{#1}%
        \expandafter\@tempa
    \fi
}
\expandafter\@tempa\colorlist,\relax,
\makeatother

\newcommand{\hhide}[1]{}

\ifx\diagnoselabel\undefined
	\relax
\else
	\makeatletter
	\def\@testdef #1#2#3{%
		\def\reserved@a{#3}\expandafter \ifx \csname #1@#2\endcsname
			\reserved@a  \else
			\typeout{^^Jlabel #2 changed:^^J%
				\meaning\reserved@a^^J%
				\expandafter\meaning\csname #1@#2\endcsname^^J}%
			\@tempswatrue \fi}
	\makeatother
\fi

\usepackage{graphicx}
\usepackage{multirow, threeparttable, booktabs, makecell}
\usepackage{amssymb}

\newcommand{\tb}[1]{\textbf{#1}}
\newcommand{\tu}[1]{\underline{#1}}
\newcommand{\tub}[1]{\underline{\textbf{#1}}}

\title{DistilVPR: Cross-Modal Knowledge Distillation for Visual Place Recognition}
\author {
    Sijie Wang\textsuperscript{\rm 1}\equalcontrib,
    Rui She\textsuperscript{\rm 1}\equalcontrib,
    Qiyu Kang\textsuperscript{\rm 1}\footnote{Corresponding author: Qiyu Kang.},
    Xingchao Jian\textsuperscript{\rm 1},
    Kai Zhao\textsuperscript{\rm 1},
    Yang Song\textsuperscript{\rm 2},
    Wee Peng Tay\textsuperscript{\rm 1}
}
\affiliations {
    \textsuperscript{\rm 1}Nanyang Technological University\\
    \textsuperscript{\rm 2}C3.AI\\
    \{wang1679, rui.she, qiyu.kang, xingchao001\}@ntu.edu.sg, yang.song@c3.ai
}

\usepackage{bibentry}

\begin{document}

\nocopyright 
\maketitle

\begin{abstract}
The utilization of multi-modal sensor data in visual place recognition (VPR) has demonstrated enhanced performance compared to single-modal counterparts. Nonetheless, integrating additional sensors comes with elevated costs and may not be feasible for systems that demand lightweight operation, thereby impacting the practical deployment of VPR. To address this issue, we resort to knowledge distillation, which empowers single-modal students to learn from cross-modal teachers without introducing additional sensors during inference. Despite the notable advancements achieved by current distillation approaches, the exploration of feature relationships remains an under-explored area. In order to tackle the challenge of cross-modal distillation in VPR, we present DistilVPR, a novel distillation pipeline for VPR. We propose leveraging feature relationships from multiple agents, including self-agents and cross-agents for teacher and student neural networks. Furthermore, we integrate various manifolds, characterized by different space curvatures for exploring feature relationships. This approach enhances the diversity of feature relationships, including Euclidean, spherical, and hyperbolic relationship modules, thereby enhancing the overall representational capacity. The experiments demonstrate that our proposed pipeline achieves state-of-the-art performance compared to other distillation baselines. We also conduct necessary ablation studies to show design effectiveness. The code is released at: 
\url{https://github.com/sijieaaa/DistilVPR}
\end{abstract}

\section{Introduction}

\begin{figure}[!t]
\begin{center}
\includegraphics[width=0.45\textwidth]{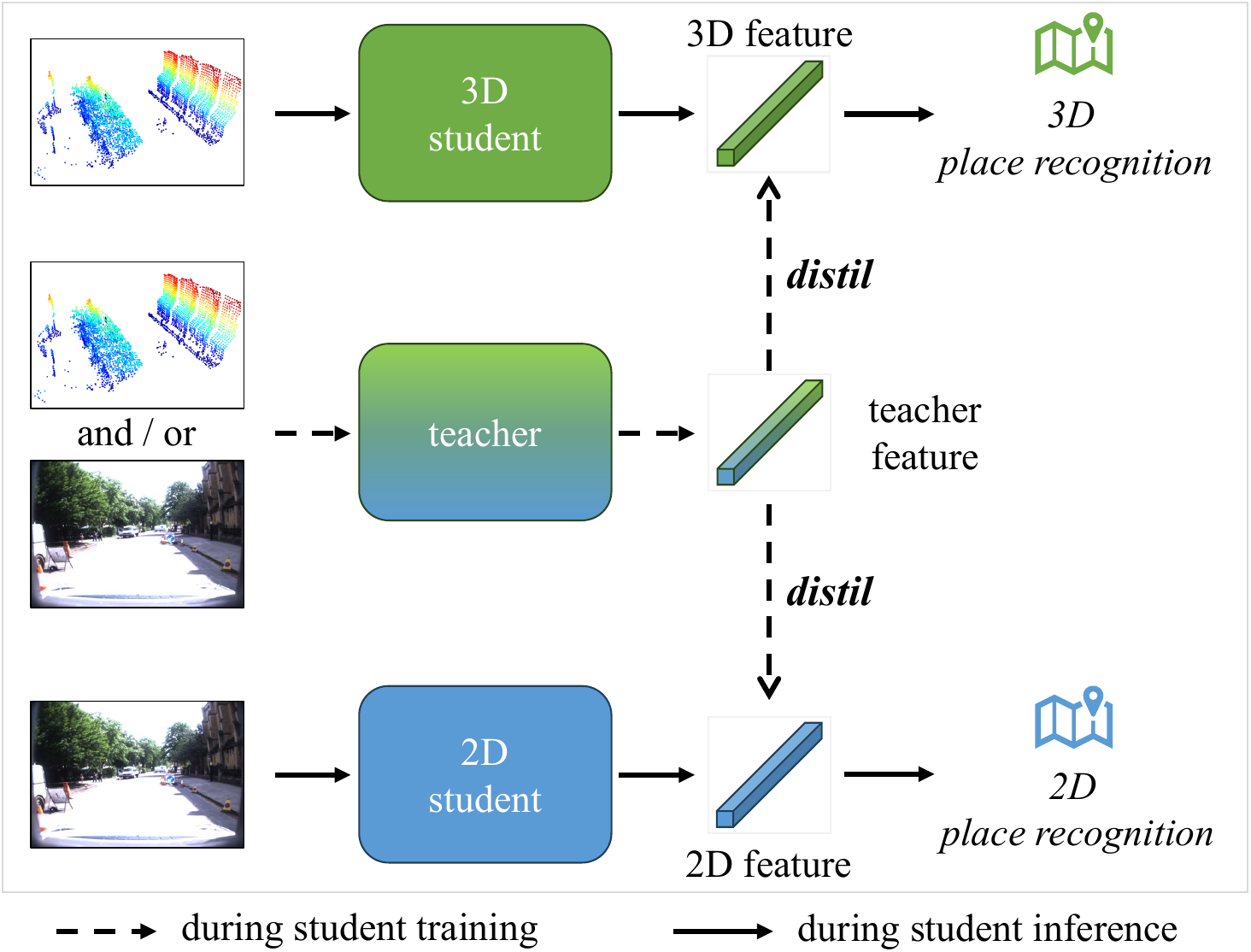}
\end{center}
\caption{The pipeline of cross-modal KD to transfer knowledge from the cross-modal teacher to single-modal students.}
\label{fig:demo}
\end{figure}

Visual place recognition (VPR) serves as a foundational task in localization, aiming at identifying locations by comparing visual sensor data, such as camera images and LiDAR point clouds, to stored references in a database. This task finds application in diverse domains, including autonomous driving \cite{chen2023deepmapping2}, precise positioning \cite{hloc_sarlin2019coarse}, and augmented reality \cite{sarlin2022lamar}.

Traditional VPR solutions rely on handcrafted features like Vector of Locally Aggregated Descriptor (VLAD) \cite{jegou2011vlad} and Bag of Words (BoW) \cite{galvez2012bagofwords}. These methods often fall short in challenging conditions including changing lighting, view distortions, and environmental perturbations due to their dependence on manual design.

The rise of deep learning has inspired data-driven VPR approaches that can tackle these challenges. NetVLAD \cite{arandjelovic2016netvlad} skillfully combines deep convolutional networks (CNNs) and traditional VLAD to enhance the robustness and efficacy of image scene feature extraction. This innovation paved the way for subsequent learning-based strategies. Addressing camera sensitivity to illumination, PointNetVLAD \cite{uy2018pointnetvlad} suggests utilizing point clouds from LiDARs which, unlike cameras, actively project laser beams to perceive surroundings, thus rendering them more resistant to lighting variations. Moreover, integrating data from multiple sensors can yield a more resilient and high-performing VPR model. In this vein, MinkLoc++ \cite{komorowski2021minkloc++} leverages both images and point clouds as inputs to achieve efficient multi-modal feature extraction, showcasing superiority over single-modal alternatives.

While integrating various sensors can elevate model performance, it also incurs additional expenses. Moreover, lightweight mobile systems might not support heavy sensors, such as LiDARs, making multi-modal sensors impractical.

Although using multiple sensors during inference is not favored, we can harness this cross-modal knowledge during student model training. This is where cross-modal knowledge distillation (KD) enters the picture. Specifically, in the student training phase, as depicted in \cref{fig:demo}, distinct modalities can be fed into the pre-trained teacher model. The extracted teacher features can then guide single-modal student models in learning superior features through additional supervision. During inference, student models can still rely on single-modal data, eliminating the need to accommodate multiple sensors.

Given the inconsistency in feature embedding across different modalities, directly compelling students to learn teacher features would be intricate. In contrast, the relational KD paradigm \cite{RKD_park2019relational}, which delves into feature relationships, offers a more suitable approach to address this inconsistency. However, the vanilla relational KD solution only considers feature relationships in limited embedding spaces, and they restrict relationship computing within the same knowledge agents (i.e. either teacher-teacher relationships or student-student relationships). These limitations hinder the efficient transfer of knowledge from teachers to students.

To mitigate these issues, we propose DistilVPR, a novel cross-modal distillation pipeline for VPR. Our contributions can be summarized as follows:
\begin{itemize}
\item
We present DistilVPR, a cross-modal KD solution uniquely tailored for VPR. This framework extends the scope of feature relationships, encompassing both \textit{self-agents} and \textit{cross-agents} to facilitate a more comprehensive exploration of knowledge. In addition, our approach performs feature embedding in \textit{multiple manifolds} with diverse feature geodesic measurements, enhancing the construction of effective feature relationships

\item Through extensive experiments, we showcase the remarkable performance of DistilVPR when compared to previous KD baselines. Our approach achieves state-of-the-art (SOTA) performance in the task of cross-modal distillation for VPR. Furthermore, we rigorously investigate our design through vital ablation studies, providing empirical evidence of the efficacy of our proposed methodology.
\end{itemize}

\section{Related Work and Preliminary}
In this section, we introduce related works and necessary manifold preliminaries.
\subsection{Visual Place Recognition}

NetVLAD \cite{arandjelovic2016netvlad} pioneers the combination of the traditional VLAD descriptor and the CNN to construct a learnable aggregation layer. Its success has paved the road for many VPR models. 
PointNetVLAD \cite{uy2018pointnetvlad} leverages point clouds instead of images to conduct place recognition. The point cloud features are extracted by PointNet \cite{qi2017pointnet} and then fed into a NetVLAD layer to produce the final global descriptor of the scene. 
MinkLoc3D \cite{komorowski2021minkloc3d} is built based on a sparse 3D CNN for point cloud feature expression. 

The aforementioned works use either images or point clouds for VPR. We now review approaches that take the multi-modal fusion strategy.
Cues-Net \cite{oertel2020cues} generates pseudo 3D point clouds from image sequences using Direct Sparse Odometry (DSO)\cite{engel2017dso}. 
PIC-Net \cite{lu2020picnet} transforms night images into the daytime style to reduce the impact of illumination perturbations on images. 
CORAL~\cite{pan2021coral} projects the 3D point cloud using bird's-eye-view (BEV) mapping, such that a 2D image backbone can be applied on both the point cloud branch and the image branch.  
MinkLoc++~\cite{komorowski2021minkloc++} follows the style of the MinkLoc3D series to achieve sparse 3D feature representation. The final global descriptor is concatenated with the 2D image descriptor and 3D point cloud descriptor.
AdaFusion~\cite{lai2022adafusion} leverages a multi-scale attention module that hierarchically aggregates multi-modal features.

\subsection{Knowledge Distillation}
KD has emerged as a pivotal technique in model compression and multi-modal learning, enabling the transfer of knowledge from complex teacher models to compact or cross-modal student models. 
Vanilla KD \cite{KD_hinton2015distilling} first introduces the concept of KD to compress the knowledge from larger teacher models to smaller student models. 
RKD \cite{RKD_park2019relational} emphasizes the self-relationships present in the data samples of both teacher and student outputs. This approach serves as an implicit distillation solution, facilitating the transfer of the teacher's knowledge to the student model. 
AFD \cite{AFD_ji2021show} employs an attention-based meta-network to acquire relative similarities among features and then employs these identified similarities to regulate the intensity of distillation for all feasible pairs. 
MKD \cite{MKD_jin2023multi} conducts prediction alignment at the instance of three different levels simultaneously, which include the instance, batch, and class levels. 

There are also some other distillation works focusing on various tasks.
2DPass \cite{20222dpass} employs an innovative approach to enhance semantic information extraction from multi-modal data with the integration of two key components: auxiliary modal fusion and multi-scale fusion-to-single distillation. 
LSD-Net \cite{lsdnet_peng2022lsdnet} leverages dual distillation to transfer teacher patterns into students for lightweight VPR.
EPC-Net \cite{epcnet_hui2022efficient} proposes ProxyConv, which is a lightweight module for local geometric feature aggregation. It uses a grouped VLAD network to form the global descriptors. To train its more lightweight version, the final feature is distilled from the larger teacher network to the smaller student network.
CSD \cite{CSD_wu2022contextual} represents a flexible framework for asymmetric similarity distillation to enhance the small query model for image retrieval.
UniDistill \cite{zhou2023unidistill} digs into BEV object detection and leverages KD from features, relationships, and responses. 
LiDAR2Map \cite{wang2023lidar2map} presents an online camera-to-LiDAR distillation scheme to facilitate semantic information from images to point clouds for semantic map segmentation.

\subsection{Manifold Preliminary}
The concept of a manifold \cite{zhao2023adversarial,wang2023hypliloc,she2023robustmat,shi2023synchrosqueezed} serves as a generalization of surfaces in higher dimensions, extending the notion of well-behaved geometrical structures. A manifold $\mathcal{M}$ is a topological space that locally resembles the Euclidean space near each point $p \in \calM$. For each point $p$, it is possible to establish a homeomorphism between a neighborhood of $p$ and the Euclidean space.

The tangent space $T_{p} \mathcal{M}$ at a point $p$ on $\mathcal{M}$ can be visualized as a hyperplane that provides the best approximation of $\mathcal{M}$ in the vicinity of $p$. Alternatively, $T_{p} \mathcal{M}$ is the space that encompasses all the possible directions of curves on $\mathcal{M}$ passing through $p$. The elements residing within $T_{p} \mathcal{M}$ are referred to as tangent vectors. Essentially, the tangent space $T_{p} \mathcal{M}$ characterizes the local linear approximation of $\mathcal{M}$ near the point $p$. It captures the intrinsic geometry of $\mathcal{M}$.

A metric tensor $g_{p}$ is an additional structure associated with each point $p$ on a manifold $\calM$. By smoothly varying across $\calM$, the metric tensor provides a consistent way to measure distances throughout the manifold.
Given two points $p, q \in \calM$, the geodesic distance $d(p, q)$ is obtained as the shortest length of curves that connect point $p$ and $q$.




\section{Proposed Pipeline}
In this section, we first provide the problem formulation. Then, we introduce the DistilVPR architecture in detail.

\subsection{Problem Formulation}

In this study, we address the challenge of cross-modal KD for VPR. We focus on a scenario where a pre-trained teacher model is provided, capable of processing images and/or point clouds as inputs for multi-modal VPR. The single-modal student models accept either image inputs or point cloud inputs. Our objective is to distill the teacher's knowledge to the students, empowering them to acquire enhanced understanding during training. This, in turn, improves student performance during inference without the requirement for cross-modal sensors.

Specifically, we denote a batch of teacher outputs as\footnote{We denote $[N]=\{1,\ldots, N \} $ for simplification.} $\bT=\set*{\bt_i\in \bbR^{C} \given i\in[N]}$  and student outputs as $\bS=\set*{\bs_i\in \bbR^{C} \given i\in[N]}$, with the batch size $N$ and the same output channel size\footnote{We assume the teacher and the student have the same output channel size.} $C$ .

\subsection{Relational Distillation}

There are typically two ways to conduct KD, including direct KD and relational KD. Direct KD is a straightforward way that directly applies sample-wise supervision by minimizing the loss
\begin{align}\label{eq:dkd}
\mathcal{L}_{\rm{direct}} = \sum_{i\in [N]} \ell\parens*{ \bt_i,  \bs_i},
\end{align}
where $\ell\parens*{}$ denotes the loss function. This approach pulls student embeddings towards teacher embeddings, which can be regarded as sample-wise supervision.

By contrast, relational KD does not apply explicit sample-wise supervision. Instead, it measures inter-sample relationships, which can be regarded as implicit knowledge. Relational KD is formed by minimizing
\begin{align}\label{eq:rkd}
\mathcal{L}_{\rm{relationship}} = \sum_{i,j\in [N]} \
\ell\parens*{ 
r\parens*{\bt_i, \bt_j},  \
r\parens*{\bs_i, \bs_j} \
},
\end{align}
where $r(\cdot, \cdot)$ is the relational function to compute embedding distances.

Through our experiments, we have observed that compared with direct KD, relational KD is inherently a better choice for cross-modal KD in VPR for the following reasons. 
On one hand, in VPR, places are recognized by computing query-database similarity, where the training goal is to minimize the query-positive distance and maximize the query-negative distance. The relative feature relationships are more critical than the absolute feature embeddings, for which the relational KD scheme that explores relative embedding distance would be a more suitable solution for VPR. 
On the other hand, cross-modal features may have inherently different embedding patterns. Thus it would be intractable to force single-modal features to be embedded in the same space as multi-modal features using direct KD schemes. Based on these insights, we follow the relational KD scheme in \cref{eq:rkd} to design a more efficient cross-modal distillation solution.

\subsection{Multi-agent Relationship}

We generically call a teacher output $\bt_i$ or student output $\bs_i$ an \emph{agent}.
One limitation of the basic relational KD is that it confines the computation of relationships within the same type of agent, i.e., teacher-teacher $r(\bt_i, \bt_j)$ and student-student $r(\bs_i, \bs_j)$. Despite relational KD being able to achieve considerably better performance than direct KD counterparts, it lacks a more generalized consideration of the combination of different agents.

\begin{figure}[!htb]
\centering
\includegraphics[width=0.38\textwidth]{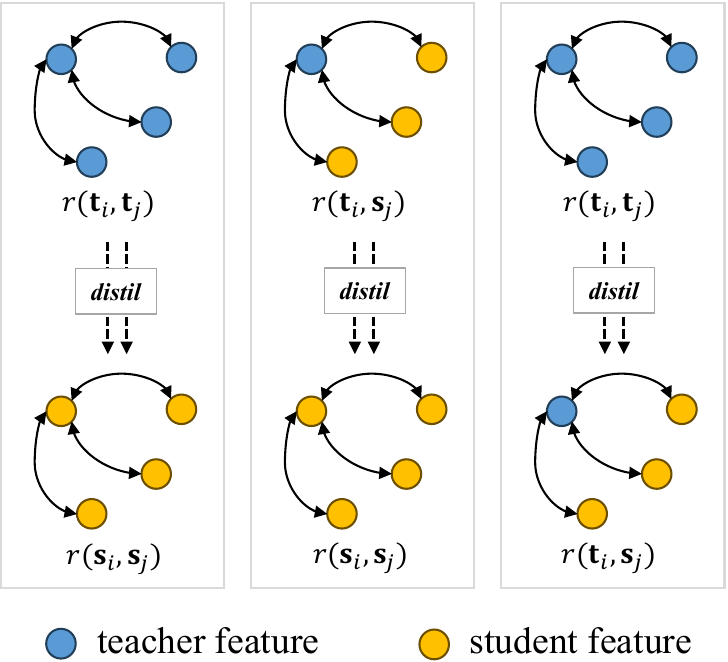}
\caption{Three generalized relational KD schemes.}
\label{fig:tt-ss}
\end{figure}

To generalize the combination of different agents, there are three scenarios for relationship computation as shown in \cref{fig:tt-ss}.
We expand \cref{eq:rkd} to further explore not only the self-agent relationships,  $r(\bt_i, \bt_j)$ and $r(\bs_i, \bs_j)$, but also the cross-agent relationship, $r(\bt_i, \bs_j)$. Specifically, the three generalized relational KD losses are formulated as:
\begin{align}
\mathcal{L}_{\rm{tt-ss}} = \sum_{i,j\in [N]} \ \label{eq:tt-ss}
\ell\parens*{ 
r\parens*{\bt_i, \bt_j},  \
r\parens*{\bs_i, \bs_j} \
},\\
\mathcal{L}_{\rm{ts-ss}} = \sum_{i,j\in [N]} \ \label{eq:ts-ss}
\ell\parens*{ 
r\parens*{\bt_i, \bs_j},  \
r\parens*{\bs_i, \bs_j} \
},\\
\mathcal{L}_{\rm{tt-ts}} = \sum_{i,j\in [N]} \ \label{eq:tt-ts}
\ell\parens*{ 
r\parens*{\bt_i, \bt_j},  \
r\parens*{\bt_i, \bs_j} \
}.
\end{align}
From above, \cref{eq:tt-ss} is the vanilla relational KD scheme in \cref{eq:rkd}. By contrast, \cref{eq:ts-ss} and \cref{eq:tt-ts} are two additional schemes with \cref{eq:tt-ts} used in CSD \cite{CSD_wu2022contextual}. In the two schemes, cross-agent relationship $r \parens*{\bt_i, \bs_j}$ bridges the gap between teacher features and student features. The additional information would contribute to a more effective KD process. 

Comparing \cref{eq:ts-ss} and \cref{eq:tt-ts}, we have empirically found that \cref{eq:ts-ss} generally outperforms \cref{eq:tt-ts}, as seen in \cref{tab:ablation}. 
This may be attributed to the fact that within the four variables in \cref{eq:tt-ts}, only one pertains to the learnable student features, while the remaining three variables are associated with the fixed teacher features. Consequently, the optimal solution domain for minimizing \cref{eq:tt-ts} becomes constrained. For example, considering $r$ as the Euclidean distance function, the optimal solution is confined to a sphere. Similarly, this constraint could potentially elucidate why direct KD in \cref{eq:dkd} yields inferior results compared to relational KD, given that direct KD also involves only one variable for student features, resulting in the optimal solution domain that is a single point.

Based on these insights, we thus use \cref{eq:ts-ss} to compute cross-agent relationships, along with \cref{eq:tt-ss} for self-agent relationships. These distinct relationship patterns constitute the core components of our approach.


\subsection{Multi-manifold Relationship}
Since different modal features may not be embedded similarly, it becomes essential to adopt a more comprehensive metric for measuring agent relationships. Consequently, we introduce combined manifold spaces to augment the effectiveness of relational KD.

Different feature manifolds can be categorized based on their curvature. The Euclidean space represents the most prevalent manifold with zero curvature, while the spherical manifold exhibits positive curvature, and the hyperbolic manifold has negative curvature. By amalgamating multiple manifolds, we can facilitate features to possess more comprehensive embedding relationships by leveraging distinct geodesic distances.

\subsubsection{Euclidean Relationship.} 
The Euclidean space serves as a prominent example of a flat manifold, exhibiting zero curvature across all points. Within Euclidean space, the calculation of the geodesic distance between any two points is given by the conventional Euclidean distance formula. The distance $d_{\rm{euc}}$ is the straight-line distance between two points $\bx, \by$ in a Cartesian coordinate system given by
\begin{align}
d_{\rm{euc}}\parens*{\bx,\by} = \norm{\bx - \by},
\end{align}
where $\norm{\cdot}$ denotes the $\calL_2$ norm. 

In our work, the Euclidean distance yields the Euclidean-based losses as:
\begin{align}
\calL^{\rm{euc}}_{\rm{tt-ss}} = \sum_{i,j\in [N]} \ell \parens*{   \
d_{\rm{euc}}\parens*{\bt_i,\bt_j},  \
d_{\rm{euc}}\parens*{\bs_i,\bs_j}   \
}, \\
\calL^{\rm{euc}}_{\rm{ts-ss}} = \sum_{i,j\in [N]} \ell \parens*{   \
d_{\rm{euc}}\parens*{\bt_i,\bs_j},  \
d_{\rm{euc}}\parens*{\bs_i,\bs_j}   \
}.
\end{align}


\subsubsection{Spherical Relationship.}

The second relationship we consider is the spherical relationship. In contrast to Euclidean space, the spherical manifold displays a distinct characteristic by possessing a constant positive curvature. The geodesic distance between two points is calculated based on the angular separation between the points and the radius of the sphere. Following previous works 
 \cite{zhou2023unidistill,pvkd_hou2022point}, we adopt the cosine distance to explore the spherical-based relationship, which is given by 
\begin{align}
d_{\rm{cos}}(  \bx,\by  ) = \frac{    \angles{\bx,\by}   }{ \norm{\bx} \norm{\by} }, 
\end{align}
where $\angles{\cdot, \cdot}$ denotes the inner product.

Then we incorporate the cosine distance as the second consideration to explore positive-curvature relationships, and the losses are formulated as: 
\begin{align}
\mathcal{L}^{\rm{cos}}_{\rm{tt-ss}} = \sum_{i,j\in [N]} \ell
\parens*{
d_{\rm{cos}}  (\bt_i,\bt_j) ,
d_{\rm{cos}}  (\bs_i,\bs_j )
}, \\
\mathcal{L}^{\rm{cos}}_{\rm{ts-ss}} = \sum_{i,j\in [N]} \ell
\parens*{
d_{\rm{cos}}  (\bt_i,\bs_j) ,
d_{\rm{cos}}  (\bs_i,\bs_j )
}.
\end{align}

\subsubsection{Hyperbolic Relationship.} 

A comprehensive relationship evaluation would benefit more from various feature pattern exploration of KD agents, and it can thus contribute to more effective KD from the teacher to the cross-modal student. However, the above two measurements explore feature relationships in either zero-curvature or positive-curvature manifolds as in RKD \cite{RKD_park2019relational}. There is a lack of consideration of relationships in negative curvature manifolds, which would result in insufficient KD. To this end, we introduce the third relationship based on the negative curvature manifold. 

In Riemannian geometry, the hyperbolic space is defined as the Riemannian manifold with constant negative curvature.  The Poincar\'e ball is the most common conformal model of hyperbolic geometry. It has been used to embed features in various tasks \cite{hyperbolic_nlp_1_tifrea2018poincar, hyperbolic_gnn_liu2019hyperbolic, wang2023hypliloc}. The $n$-dimensional Poincar\'e ball is defined on $\bbD_c^n= \set*{\bp \in \bbR^n \given  c\norm{\bp}<1}$ with curvature  $-c^2$. The Poincar\'e ball is equipped with a constant metric tensor $\bg=\lambda^2_c \bI^n$, where  $\lambda_c = \frac{2}{1-c\norm{\bp}^2}$ is the conformal factor.

Given a pair $\bp, \bq \in \bbD_c^n$, the mobius addition $\oplus_c$ is defined as:
\begin{align}
\bp \oplus_c \bq=\frac{\parens*{1+2c\angles{\bp,\bq} + c \norm{\bq}^2}\bp + \parens*{1-c\norm{\bp}^2}\bq}{1+2c\angles{\bp, \bq} + c^2\norm{\bp}^2 \norm{\bq}^2 }.
\end{align}

For a fixed base point $\bz \in \mathbb{D}_c^n$, the exponential mapping function $\exp_{\bz}^c: \mathbb{R}^n \rightarrow \mathbb{D}_c^n$ maps points from the tangent Euclidean space to the hyperbolic space:
\begin{align}
\exp_{\bz}^c(\bv)=\bz \oplus_c   \parens*{\tanh    \parens*{\sqrt{c} \frac{\lambda_c \norm{\bv} }{2}}   \frac{\bv}{\sqrt{c} \norm{\bv} }  }.
\end{align}

By setting the origin as the fixed base point, the exponential map can be simplified as
\begin{align}
\exp^c_{0} (\bv) = \tanh\parens*{\sqrt{c} \norm{\bv}}    \frac{\bv}{\sqrt{c}\norm{\bv}}.
\end{align}

After exponential mapping, the geodesic distance in the hyperbolic manifold (hyperbolic distance) can be obtained as 
\begin{align}
d_{\rm{hyp }}(\bp, \bq)=\frac{2}{\sqrt{c}} \operatorname{arctanh}   \parens*{ \sqrt{c} \norm{- \bp \oplus_c \bq} }.
\end{align}

In our work, we embed both teacher outputs and student outputs in the Poincar\'e ball, and the hyperbolic losses are computed as:
\begin{align}
\calL_{\rm{tt-ss}}^{\rm{hyp}} = \sum_{i,j\in [N]} \ell \parens*{   \
d_{\rm{hyp}}\parens*{\bt_i^{\rm{hyp}},\bt_j^{\rm{hyp}}}, d_{\rm{hyp}}\parens*{\bs_i^{\rm{hyp}},\bs_j^{\rm{hyp}}}   \
}, \\
\calL_{\rm{ts-ss}}^{\rm{hyp}} = \sum_{i,j\in [N]} \ell \parens*{   \
d_{\rm{hyp}}\parens*{\bt_i^{\rm{hyp}},\bs_j^{\rm{hyp}}}, d_{\rm{hyp}}\parens*{\bs_i^{\rm{hyp}},\bs_j^{\rm{hyp}}}   \
}, 
\end{align}
where $\bt_i^{\rm{hyp}} = \rm{exp}^c_0(\bt_i)$ and  $\bs_i^{\rm{hyp}} = \rm{exp}^c_0(\bs_i)$ are hyperbolic  teacher and student embeddings, respectively.

\subsection{Overall Loss Function}
Finally, we combine the insights from multiple agents and multiple manifolds to construct our distillation pipeline. Specifically, we first formulate two distillation losses, including the self-agent distillation loss $\calL_{\rm{KD-S}}$ and the cross-agent distillation loss $\calL_{\rm{KD-C}}$ respectively as:
\begin{align}
\calL_{\rm{KD-S}} =  \calL_{\rm{tt-ss}}^{\rm{euc}} + \calL_{\rm{tt-ss}}^{\rm{cos}} + \calL_{\rm{tt-ss}}^{\rm{hyp}},
\end{align}
\begin{align}
\calL_{\rm{KD-C}} =  \calL_{\rm{ts-ss}}^{\rm{euc}} + \calL_{\rm{ts-ss}}^{\rm{cos}} + \calL_{\rm{ts-ss}}^{\rm{hyp}}.
\end{align}
Subsequently, with weight hyperparameters $\lambda_{\rm{S}}, \lambda_{\rm{C}}$ and the triplet loss as the VPR task loss $\calL_{\rm{task}}$, we propose three different overall losses. They are denoted as DistilVPR-S, DistilVPR-C, and DistilVPR-SC, respectively:
\begin{align}\label{eq:s}
\calL_{\rm{DistilVPR-S}} = \calL_{\rm{task}} + \lambda_{\rm{S}}\calL_{\rm{KD-S}},
\end{align}
\begin{align}\label{eq:c}
\calL_{\rm{DistilVPR-C}} = \calL_{\rm{task}} + \lambda_{\rm{C}}\calL_{\rm{KD-C}},
\end{align}
\begin{align}\label{eq:sc}
\calL_{\rm{DistilVPR-SC}} = \calL_{\rm{task}} + \lambda_{\rm{S}}\calL_{\rm{KD-S}} +  \lambda_{\rm{C}}\calL_{\rm{KD-C}}.
\end{align}

\section{Experiments}
In this section, we conduct experiments to compare DistilVPR defined in \cref{eq:s,eq:c,eq:sc} with other KD baselines. We also provide necessary ablation studies to verify the design efficacy.

\subsection{Datasets and Implementation Details}
\subsubsection{Oxford RobotCar.}
The Oxford RobotCar dataset \cite{oxford} is a large-scale autonomous driving dataset, which provides a rich collection of sensor data, including images and point clouds. It also encompasses various driving scenarios with different weather conditions, traffic patterns, and pedestrian interactions. We use the processed point clouds provided by PointNetVLAD\cite{uy2018pointnetvlad} which is the standard benchmark data for point cloud and multi-modal (image + point cloud) place recognition. Since it is equipped with both images and point clouds, the Oxford RobotCar dataset would be a suitable platform to test the performance of multi-modal teachers and single-modal students.

\subsubsection{Boreas.}
The Boreas dataset \cite{burnett2022boreas} is gathered by conducting multiple drives along a consistent route over one year, thereby capturing notable seasonal fluctuations. It comprises an extensive collection of over 350 km of driving data, featuring numerous sequences recorded under challenging weather conditions, including rain, heavy snow, and night. It also provides multi-modal sensor data such as images and point clouds, and thus can also serve as a benchmark for both multi-modal and single-modal models.

\subsubsection{Implementation Details.}
We choose two SOTA multi-modal place recognition models as teachers, including MinkLoc++ \cite{komorowski2021minkloc++} and AdaFusion \cite{lai2022adafusion}. We use their single-modal branches as students to test the effectiveness of cross-modal KD. We use the Adam optimizer to train both teachers and students. The learning rate is set as $1e-4$ and $1e-3$ for the image branch and the point cloud branch respectively. Both teacher models and student models are trained for $60$ epochs with $128$ batch size. All experiments are conducted on an A100 GPU. We follow previous works to use the same evaluation protocol, including Average Recall@1 (AR@1) and Average Recall@1\% (AR@1\%). More details are provided in the supplement.

\subsection{Main Results} 

\subsubsection{Fusion-to-single Distillation.}
As shown in \cref{tab:oxford} and \cref{tab:boreas}, our proposed three KD schemes can achieve considerably better performance compared with other counterparts in the Oxford and the Boreas datasets. In addition, our schemes can handle various fusion-to-single KD tasks, including fusion-to-2D and fusion-to-3D, which further underscores the efficacy and generalization ability. We have also noticed that relational KD schemes generally outperform the direct KD counterparts, which shows that the key to effective distillation for VPR lies in the exploration of feature relationships rather than mere feature alignment.

Moreover, we have found that the 3D point cloud inputs can always contribute better VPR performance compared with the 2D image inputs. This trend holds across both datasets, with the gap being particularly pronounced in the more challenging Boreas dataset. This observation reinforces the assertion that utilizing point cloud data is pivotal in achieving effective VPR results.

\begin{table*}[!htb]\footnotesize
\centering
\begin{tabular}{l | c c   c c    c c   c  c }
\toprule
\multirow{3}{*}{Distillation Method}  &  \multicolumn{2}{c}{T: MinkLoc++}   &  \multicolumn{2}{c}{T: MinkLoc++}  & \multicolumn{2}{c}{T: AdaFusion}   &  \multicolumn{2}{c}{T: AdaFusion}  \\
&  \multicolumn{2}{c}{S: MinkLoc++2D} &  \multicolumn{2}{c}{S: MinkLoc++3D}  &   \multicolumn{2}{c}{S: AdaFusion-2D}  &   \multicolumn{2}{c}{S: AdaFusion-3D}  \\
&  AR@1\% & AR@1  &  AR@1\% & AR@1  &  AR@1\% &  AR@1   &  AR@1\% &  AR@1  \\
\midrule
Teacher             &  99.4  &  97.2  &  99.4  &  97.2 &  99.0  &  96.6  & 99.0 &  96.6 \\
Student w/o distil. &  94.7  &  85.7  &  98.1  &  94.4 &  94.2  &  84.2  & 98.0 &  93.8 \\
\midrule
\midrule
*KD \cite{KD_hinton2015distilling}   &  95.2  &  84.6  &  98.1  &  94.3  &  95.2  &  84.6  & 97.8 &  93.7 \\
*AFD \cite{AFD_ji2021show} &  95.2  &  84.7  &  98.1  &   94.2 &  94.5  &  82.9  & 97.8 &  93.2 \\
*EPC-Net \cite{epcnet_hui2022efficient} &  95.4  &  85.6  &  97.9  &   93.8  &  95.3  &  85.1  & 98.1 &  93.7 \\
\midrule
\midrule
RKD \cite{RKD_park2019relational}    &  96.5  &  88.5  &  \tub{98.3}  &   94.5  &  96.2  &  87.3  & \tu{98.2} &  \tu{94.4} \\
CSD \cite{CSD_wu2022contextual}      &  95.4  &  86.0  &  98.1  &   94.4  &  95.3  &  85.4  & 98.0 &  93.8 \\
LSD-Net \cite{lsdnet_peng2022lsdnet}   &  95.8  &  86.1  &  98.2  &   94.1 &  95.9  &  86.2  & 97.9 &  93.9 \\
MKD \cite{MKD_jin2023multi}   &  95.0  &  85.4  &  98.1  &   93.9 &  95.1  &  84.5  & 97.8 &  93.7 \\
\midrule
 (ours) DistilVPR-S &  96.7  &  88.7 &  \tub{98.3}  &   \tub{95.2}    &  96.2 &  87.4  & 98.0 &  94.2 \\
 (ours) DistilVPR-C &  \tub{97.3}  &  \tub{91.1}  &  98.1  &   94.4    &  \tu{96.6} &  \tu{88.8}  & 98.0 &  93.7 \\
 (ours) DistilVPR-SC &  \tu{97.0}  &  \tu{90.0}  &  \tub{98.3}  &   \tu{94.6}    &  \tub{96.7} &  \tub{89.0}  & \tub{98.3} &  \tub{94.7} \\
\bottomrule
\end{tabular}
\caption{Fusion-to-single distillation comparison on the Oxford RobotCar dataset. "T:" and "S:" stand for the teacher model and the student model respectively. Direct distillation solutions are marked with "*", while relational solutions are without any mark. The best results are bold and underlined, while the second-best results are underlined only. }
\label{tab:oxford}
\end{table*}

\begin{table*}[!htb]\footnotesize
\centering
\begin{tabular}{l  | c c   c c    c c   c  c }
\toprule
\multirow{3}{*}{Distillation Method}  &  \multicolumn{2}{c}{T: MinkLoc++}   &  \multicolumn{2}{c}{T: MinkLoc++}  & \multicolumn{2}{c}{T: AdaFusion}   &  \multicolumn{2}{c}{T: AdaFusion}  \\
&  \multicolumn{2}{c}{S: MinkLoc++2D} &  \multicolumn{2}{c}{S: MinkLoc++3D}  &   \multicolumn{2}{c}{S: AdaFusion-2D}  &   \multicolumn{2}{c}{S: AdaFusion-3D}  \\
&  AR@1\% & AR@1  &  AR@1\% & AR@1  &  AR@1\% &  AR@1   &  AR@1\% &  AR@1  \\
\midrule
Teacher             &  98.9  &  93.1  &  98.9  &   93.1 &  98.9  &  93.2  & 98.9 &  93.2 \\ 
Student w/o distil. &  75.2  &  60.0  &  98.5  &   91.0 &  74.5  &  59.6  & 98.9 &  91.5 \\
\midrule
\midrule
*KD \cite{KD_hinton2015distilling}   &  75.8  &  61.4  &  98.1  &   90.4 &  76.9  &  60.3  & 98.5 &  91.7 \\
*AFD \cite{AFD_ji2021show} &  75.5  &  60.4  &  97.4  &  88.4 &  75.9  &  58.5  & 98.7 &  92.7 \\
*EPC-Net \cite{epcnet_hui2022efficient} &  75.3  &  60.8  &  98.0  &   89.8 &  75.4  &  60.5  & 98.9 &  92.4 \\
\midrule
\midrule
RKD \cite{RKD_park2019relational}    &  \tu{78.0}  &  62.9  &  \tub{99.1}  &   91.6 &  78.8  &  62.5  & 98.9 &  \tu{93.9} \\
CSD \cite{CSD_wu2022contextual}        &  76.3  &  61.4  &  98.2  &   90.9  &  77.0  &  61.2  & 99.1 &  92.8 \\
LSD-Net \cite{lsdnet_peng2022lsdnet}   &  74.5  &  59.3  &  \tu{98.7}  &   \tu{92.0} &  76.7  &  60.4  & 98.5 &  92.2 \\
MKD \cite{MKD_jin2023multi}            &  77.6  &  61.3  &  96.9  &   88.2 &  76.5  &  61.0  & 98.9 &  92.2 \\
\midrule
(ours) DistilVPR-S        &  77.3  &  63.4  &  98.5  &  \tub{92.1}    &  \tub{80.1} &  64.1  & \tub{99.3} &  \tub{94.0} \\
(ours) DistilVPR-C       &  77.0  &  \tu{65.1}  &  97.8  &  90.5    &  \tu{79.7} &  \tu{64.7}  & 98.8 &  92.3 \\
(ours) DistilVPR-SC      &  \tub{79.3}  &  \tub{67.2}  &  98.3  &  91.3    &  78.0 &  \tub{65.5}  & \tub{99.3} &  93.6 \\
\bottomrule
\end{tabular}
\caption{Fusion-to-single distillation comparison on the Boreas dataset.}
\label{tab:boreas}
\end{table*}

\subsubsection{3D-to-2D and Big-to-small Distillation.}
We evaluate the cross-modal distillation performance by training teachers with pure 3D point cloud inputs and students with pure 2D images. As illustrated in \cref{tab:3d-to-2d}, the distinct advantages of DistilVPR become more evident in this context. Notably, in the 3D-to-2D scenarios, DistilVPR-SC exhibits notably superior performance compared to other baselines. This result underscores the pronounced effectiveness of our methodology in addressing the intricate challenge of distillation across disparate modalities. We also assess the basic scenario of distillation from a larger model to a smaller one, as presented in \cref{tab:3d-to-2d}. In this setting, our proposed approach continues to demonstrate effective distillation performance.

\begin{table*}[!htb]\footnotesize
\centering
\begin{tabular}{l  | c c   c c    c c   c  c }
\toprule
\multirow{3}{*}{Distillation Method}  &  \multicolumn{2}{c}{T: MinkLoc++2D-Big}   &  \multicolumn{2}{c}{T: MinkLoc++3D}  & \multicolumn{2}{c}{T: AdaFusion-2D-Big}   &  \multicolumn{2}{c}{T: AdaFusion-3D}  \\
&  \multicolumn{2}{c}{S: MinkLoc++2D} &  \multicolumn{2}{c}{S: MinkLoc++2D}  &   \multicolumn{2}{c}{S: AdaFusion-2D}  &   \multicolumn{2}{c}{S: AdaFusion-2D}  \\
&  AR@1\% & AR@1  &  AR@1\% & AR@1  &  AR@1\% &  AR@1   &  AR@1\% &  AR@1  \\
\midrule
Teacher             &  80.3  &  66.4  &  98.5  &   91.0  &  87.8  &  64.7  & 98.9  &  91.5 \\ 
Student w/o distil. &  75.2  &  60.0  &  75.2  &   60.0  &  74.5  &  59.6  & 74.5  &  59.6 \\ 
\midrule
\midrule
*KD \cite{KD_hinton2015distilling}      &  75.1  &  60.1  &  74.9  &   58.6  &  76.8  &  60.6  & 75.8 &  59.0 \\ 
*AFD \cite{AFD_ji2021show}              &  77.3  &  62.5  &  75.3  &   56.5  &  76.2  &  58.5  & 76.5 &  59.4 \\ 
*EPC-Net \cite{epcnet_hui2022efficient} &  74.8  &  59.2  &  73.5  &   58.6  &  \tub{77.3}  &  60.4  & 75.2 &  60.5 \\ 
\midrule
\midrule
RKD \cite{RKD_park2019relational}      &  76.4  &  61.7  &  76.2  &   60.4  &  75.8  &  61.5  & 77.4 &  61.4 \\ 
CSD \cite{CSD_wu2022contextual}         &  77.3  &  60.3 &  76.2  &   60.3  &  76.1  &  60.2  & 76.8 &  61.0 \\
LSD-Net \cite{lsdnet_peng2022lsdnet}   &  \tu{77.6}  &  61.9  &  75.1  &   57.0  &  74.8  &  60.3  & 74.7 &  59.1 \\ 
MKD \cite{MKD_jin2023multi}            &  77.2  &  60.1  &  75.5  &   59.0  &  74.1  &  60.0  & 75.5 &  60.1 \\ 
\midrule
(ours) DistilVPR-S       &  \tub{77.9}  &  62.0  &  76.4  &   61.3  &  76.8  &  62.2  &  77.1 &  62.6 \\ 
(ours) DistilVPR-C       &  77.0    &  \tu{64.2}   &  \tu{78.0}  &   \tu{66.4}   &  \tub{77.3}    &  \tu{64.2}  &  \tu{78.4} &  \tu{65.9} \\ 
(ours) DistilVPR-SC      &  77.1  &  \tub{65.4}  &  \tub{81.1}  &   \tub{68.2}  &  76.8  &  \tub{64.7}  &  \tub{79.0} &  \tub{66.5} \\ 
\bottomrule
\end{tabular}
\caption{Big-to-small and 3D-to-2D distillation comparison on the Boreas dataset.}
\label{tab:3d-to-2d}
\end{table*}

\subsection{Ablation Study}
\subsubsection{Agent Relationships.}
We compare the performance of different relationships as in \cref{tab:ablation}. The combination of using both self-agent and cross-agent relationships achieves optimal performance, which verifies the effectiveness of our multi-agent relationships.

\begin{table}[!htb]\footnotesize
\centering
\begin{tabular}{l    c  c  c  c  c}
\toprule
Method  &  AR@1\%  &  AR@1\\
\midrule 
w/o distil.  &  75.2  &  59.3  \\
$\calL_{\rm{tt-ss} }$ in \cref{eq:tt-ss} &  76.4  &  61.3  \\
$\calL_{\rm{ts-ss} }$ in \cref{eq:ts-ss} &  78.0  &  66.4  \\
$\calL_{\rm{tt-ts} }$ in \cref{eq:tt-ts} &  76.1  &  60.6  \\
\midrule
$\calL_{\rm{tt-ss} }$ + $\calL_{\rm{ts-ss} }$ &  \tb{81.1}  &  \tb{68.2}  \\
\bottomrule
\end{tabular}
\caption{Ablation study on the self-agent and cross-agent relationship computation.}
\label{tab:ablation}
\end{table}

\subsubsection{Manifold Relationships.}
We proceed to examine the utilization of different relationship distances, as detailed in \cref{tab:distance}. Notably, the three fundamental distances yield comparable individual performances. Further using only two manifold distances with insufficient curvature exploration could not always bring improvements compared with using a single manifold. By contrast, through the fusion of sufficient relationship distances across multiple manifolds with consideration of all types of curvatures, a remarkable enhancement in distillation performance is observed. This substantiates the effectiveness of our approach in exploiting feature relationships within diverse curvature manifolds.

\begin{table}[!htb]\footnotesize
\centering
\begin{tabular}{l l l  c  c  c}
\toprule
$d_{\rm{euc}}$  &  $d_{\rm{cos}}$  & $d_{\rm{hyp}}$    &  Ours-S  &  Ours-C  &  Ours-SC \\
\midrule
\checkmark  &  &    &   59.9  &  65.2 &  66.8\\
  & \checkmark  &   &   60.2  &  64.9 &  66.5\\
  &   & \checkmark  &   60.0  &  65.6 &  67.0\\
\midrule
\checkmark  & \checkmark &               &  60.5 &  65.8  &  67.4 \\
\checkmark  &  &  \checkmark             &  60.2 &  66.0  &  66.8 \\
  & \checkmark &  \checkmark             &  60.1 &  66.1  &  66.9 \\
\midrule
\checkmark  & \checkmark  & \checkmark   &  \tb{61.3} &  \tb{66.4}  &  \tb{68.2} \\
\bottomrule
\end{tabular}
\caption{AR@1 comparison on different distance functions and relationship agent combinations. 
}
\label{tab:distance}
\end{table}

\subsubsection{Different Teacher Modalities.}
In \cref{tab:teacher}, we present a comparison of the distillation performance achieved with different teachers. Intriguingly, it is observed that the 3D-based model MinkLoc++3D can even outperform the fusion model MinkLoc++ in terms of distillation efficiency. This finding underscores the notion that a good task performer might not necessarily translate into a good teacher for distillation.

\begin{table}[!htb]\footnotesize
\centering
\begin{tabular}{l c c c}
\toprule
Teacher & T: AR@1 &  S: AR@1 \\
\midrule
MinkLoc++        & \tb{93.1}  &  67.2   & \\
MinkLoc++3D      & 91.3  &  \tb{68.2}   & \\
MinkLoc++2D-big  & 66.4  &  65.4   & \\
\bottomrule
\end{tabular}
\caption{Distillation from different teachers. The student is MinkLoc++2D with DistilVPR-SC.}
\label{tab:teacher}
\end{table}

\subsubsection{Visualization.}
\begin{figure}[!htb]
\centering
\includegraphics[width=0.42\textwidth]{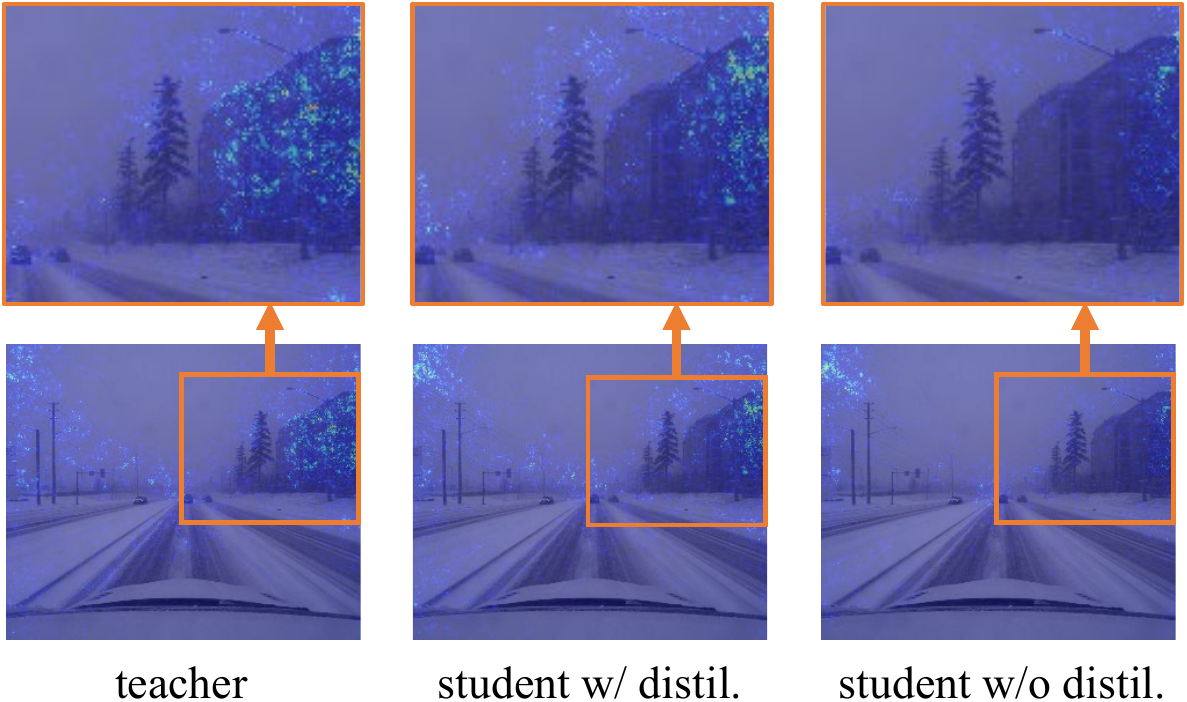}
\caption{Visualization of the salience maps. With distillation from the teacher, the student is guided to focus on scene-specific objects such as buildings.}
\label{fig:salience}
\end{figure}

A more detailed example is illustrated in \cref{fig:salience} with visualized salience maps. Distillation facilitates the student in emphasizing scene-specific objects such as buildings, which showcases the effectiveness of teacher knowledge.


\section{Conclusion and Limitations}
This paper presents DistilVPR, a novel cross-modal distillation pipeline designed for enhancing visual place recognition. We harness multi-agent and multi-manifold relationships to facilitate knowledge exploration, leading to superior performance compared to other distillation baselines.

A limitation of our approach lies in its assumption of identical feature dimensions between teachers and students, potentially restricting its applicability. Nevertheless, this limitation could be addressed by employing a feature adaptor to align the feature dimensions of teachers and students.

\section{Acknowledgement}
This research is supported by the Singapore Ministry of Education Academic Research Fund Tier 2 grant MOE-T2EP20220-0002, and the National Research Foundation, Singapore and Infocomm Media Development Authority under its Future Communications Research and Development Programme. 
The computational work for this article was partially performed on resources of the National Supercomputing Centre, Singapore (https://www.nscc.sg).

\bibliography{aaai24}

\clearpage
\newpage

\begin{center}
  {
  \large
  \lineskip .5em
  \begin{tabular}[t]{c}
        {\LARGE\bf Supplement \par}
  \end{tabular}
  \par
  }
  \vskip .5em
  \vspace*{12pt}
\end{center}

\section{Implementation Details}
We choose two SOTA multi-modal place recognition models as teachers, including MinkLoc++ \cite{komorowski2021minkloc++} and AdaFusion \cite{lai2022adafusion}. We use their single-modal branches as students to test the effectiveness of cross-modal KD. The scene descriptor dim of the MinkLoc++ series is $384$, while the scene descriptor dim of the AdaFusion series is $256$. We use the Adam optimizer to train both teachers and students. The learning rate is set as $1e-4$ and $1e-3$ for the image branch and the point cloud branch respectively. The learning rate decays by $\times0.1$ at epoch $40$. Both teacher models and student models are trained for $60$ epochs with $128$ batch size. All experiments are conducted on an A100 GPU using PyTorch. We follow previous works to use the same evaluation protocol, including Average Recall@1 (AR@1) and Average Recall@1\% (AR@1\%).

\section{More Results}

\subsection{Cross Model Distillation}
Although we assume the same output channel sizes for both teacher and student models in the main paper, we have also conducted evaluations in the cross-model scenario, where teacher and student models possess different channel sizes. To mitigate the resultant feature inconsistency, we integrate a fully-connected (fc) layer as an adaptor to harmonize feature channels.

More specifically, we examine two distinct scenarios: one with the adaptor integrated into the teacher model (as shown in \cref{stab:adaptor_tea}) and the other with the adaptor in the student model (as shown in \cref{stab:adaptor_stu}). Remarkably, our proposed DistilVPR-SC achieves SOTA performance in 7 out of 8 metrics in the two tables, thereby further validating the efficacy and broad applicability of our approach. Furthermore, when comparing the two methods of implementing the adaptor, it is generally observed that integrating the adaptor within the teacher models yields better results compared to the alternative of placing it within the student models.

\begin{table*}[!htb]\footnotesize
\centering
\begin{tabular}{l  | cc cc cc cc}
\toprule
\multirow{3}{*}{Distillation Method}  &  \multicolumn{2}{c}{T: MinkLoc++3D}   &  \multicolumn{2}{c}{T: AdaFusion-3D}  \\
&  \multicolumn{2}{c}{S: AdaFusion-2D} &  \multicolumn{2}{c}{S: MinkLoc++2D} \\
&  AR@1\% & AR@1  &  AR@1\% & AR@1  \\
\midrule
Teacher             &  98.5  &  91.0  &  98.9  &   91.5  \\ 
Student w/o distil. &  74.5  &  59.6  &  75.2  &   60.0  \\ 
\midrule
\midrule
*KD \cite{KD_hinton2015distilling}      &  76.4  &  58.9  &  74.7  &   59.9  \\ 
*AFD \cite{AFD_ji2021show}              &  73.8  &  56.9  &  74.2  &   56.9  \\ 
*EPC-Net \cite{epcnet_hui2022efficient} &  75.6  &  56.9  &  75.9  &   58.3  \\ 
\midrule
\midrule
RKD \cite{RKD_park2019relational}      &  76.3  &  60.1  &  \tu{79.0}  &   62.8  \\ 
CSD \cite{CSD_wu2022contextual}        &  74.7  &  59.9  &  76.2  &   61.2  \\ 
LSD-Net \cite{lsdnet_peng2022lsdnet}   &  74.5  &  58.2  &  74.7  &   59.4  \\ 
MKD \cite{MKD_jin2023multi}            &  \tu{76.9}  &  58.8  &  76.7  &   60.8  \\ 
\midrule
(ours) DistilVPR-S       &  76.5  &  60.3  &  78.1  &   \tu{63.2}  \\ 
(ours) DistilVPR-C       &  75.8  &  \tu{61.6}  &  77.5  &   62.1  \\ 
(ours) DistilVPR-SC      &  \tub{77.6}  &  \tub{63.1}  &  \tub{80.7}  &   \tub{66.3}  \\ 
\bottomrule
\end{tabular}
\caption{Cross model 3D-to-2D distillation comparison on the Boreas dataset. The adaptor is equipped on the teacher model.}
\label{stab:adaptor_tea}
\end{table*}

\begin{table*}[!htb]\footnotesize
\centering
\begin{tabular}{l  | c c   c c    }
\toprule
\multirow{3}{*}{Distillation Method}  &  \multicolumn{2}{c}{T: MinkLoc++3D}   &  \multicolumn{2}{c}{T: AdaFusion-3D}  \\
&  \multicolumn{2}{c}{S: AdaFusion-2D} &  \multicolumn{2}{c}{S: MinkLoc++2D} \\
&  AR@1\% & AR@1  &  AR@1\% & AR@1  \\
\midrule
Teacher             &  98.5  &  91.0  &  98.9  &   91.5  \\ 
Student w/o distil. &  74.5  &  59.6  &  75.2  &   60.0  \\ 
\midrule
\midrule
*KD \cite{KD_hinton2015distilling}      &  74.2  &  58.6  &  \tub{77.1}  &   61.1  \\ 
*AFD \cite{AFD_ji2021show}              &  76.1  &  59.6  &  75.7  &   58.8  \\ 
*EPC-Net \cite{epcnet_hui2022efficient} &  75.4  &  60.5  &  75.4  &   60.6  \\ 
\midrule
\midrule
RKD \cite{RKD_park2019relational}      &  75.9  &  \tu{62.0}  &  75.1  &   60.4  \\ 
CSD \cite{CSD_wu2022contextual}        &  74.1  &  60.2  &  \tu{76.2}  &   60.9  \\ 
LSD-Net \cite{lsdnet_peng2022lsdnet}   &  71.6  &  58.5  &  74.9  &   59.2  \\ 
MKD \cite{MKD_jin2023multi}            &  73.4  &  58.7  &  76.0  &   61.2  \\ 
\midrule
(ours) DistilVPR-S       &  \tu{76.6}  &  61.6  &  74.7  &   \tu{61.7}    \\ 
(ours) DistilVPR-C       &  \tu{76.6}  &  61.7  &  75.8  &   60.6    \\ 
(ours) DistilVPR-SC      &  \tub{77.4}  &  \tub{62.4}  &  76.1  &   \tub{63.4}    \\ 
\bottomrule
\end{tabular}
\caption{Cross model 3D-to-2D distillation comparison on the Boreas dataset. The adaptor is equipped on the student model.}
\label{stab:adaptor_stu}
\end{table*}

\subsection{Visualization}
\begin{figure}[!htb]
\centering
\includegraphics[width=0.48\textwidth]{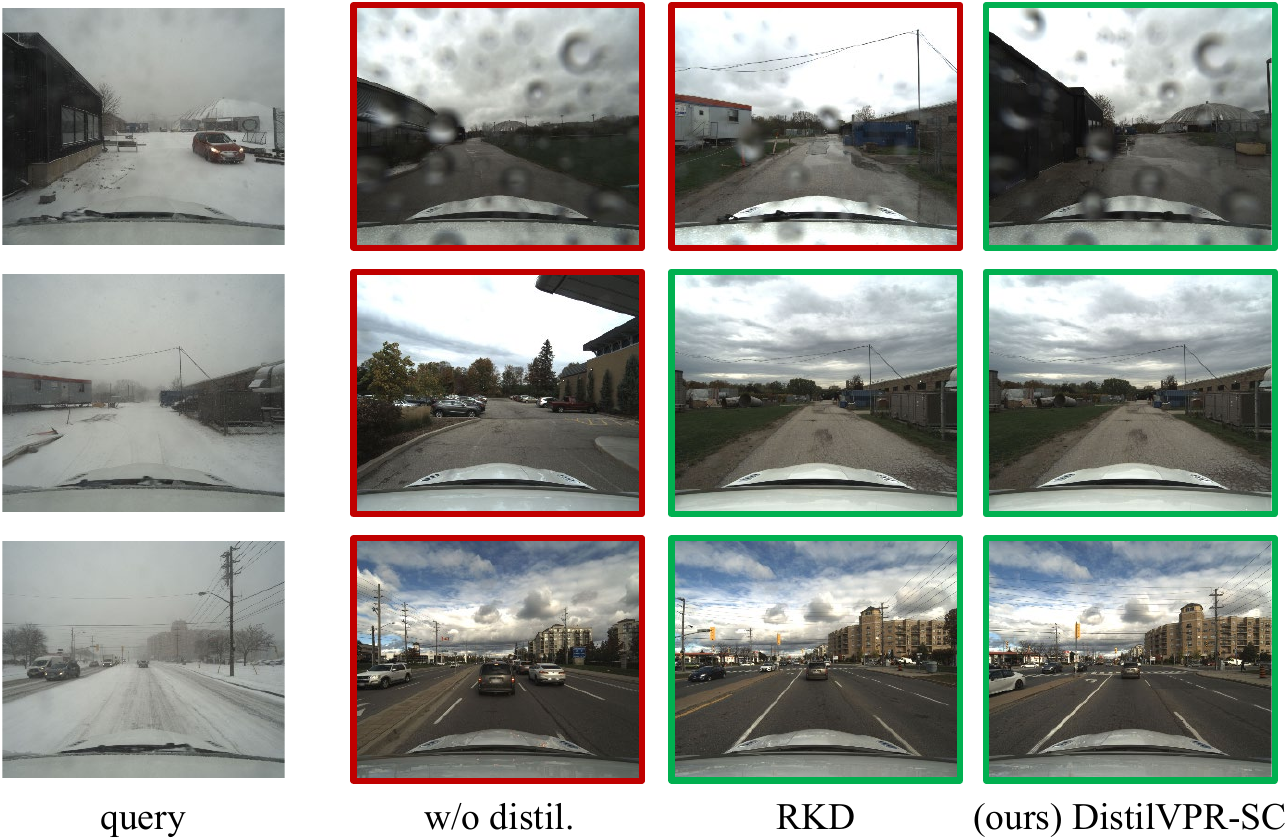}
\caption{Visualizations of place recognition examples.}
\label{sfig:viz}
\end{figure}

We present visualizations of several place recognition examples in \cref{sfig:viz}. These examples showcase the effectiveness of the distillation approaches in enhancing performance. Notably, in comparison to the results obtained using RKD \cite{RKD_park2019relational}, our proposed DistilVPR-SC yields better outcomes, particularly evident in the accurate recognition of challenging rain scenes.

\section{Compared Baselines}
The baselines we used in our experiments can be classified into direct KD solutions and relational KD solutions. 
\begin{itemize}
\item \textbf{Direct KD} baselines include: vanilla KD \cite{KD_hinton2015distilling}, AFD \cite{AFD_ji2021show}, EPC-Net \cite{epcnet_hui2022efficient}. 
\item \textbf{Relational KD} baselines include: RKD \cite{RKD_park2019relational}, CSD \cite{CSD_wu2022contextual}, LSD-Net \cite{lsdnet_peng2022lsdnet}, MKD \cite{MKD_jin2023multi}.
\end{itemize}

\section{Open Sources}
The two used datasets are both open-source datasets and can be available online.
\begin{itemize}
\item The Oxford RobotCar dataset. \\ \url{https://robotcar-dataset.robots.ox.ac.uk/}
\item The Boreas dataset. \\ \url{https://www.boreas.utias.utoronto.ca/#/}
\end{itemize}
Our code is mainly built based on the following repositories.
\begin{itemize}
\item MinkLoc++ \\ \url{https://github.com/jac99/MinkLocMultimodal}
\item AdaFusion \\ \url{https://github.com/MetaSLAM/AdaFusion}
\end{itemize}

\end{document}